\documentclass[runningheads]{llncs}
\usepackage[T1]{fontenc}
\usepackage{graphicx}
\usepackage{multirow}
\usepackage{hyperref}
\begin{document}

%%%% Title
\title{Mask-Free Neuron Concept Annotation for Interpreting Neural Networks in Medical Domain}
\titlerunning{ }

\author{Hyeon Bae Kim$^{\dag}$,
Yong Hyun Ahn$^{\dag}$,
Seong Tae Kim\textsuperscript{${\ast}$}} 
%index{Hyeon Bae, Kim}
%index{Yong Hyun, Ahn}
%index{Seong Tae, Kim}

\authorrunning{Kim, Ahn et al.}
% \authorrunning{}

\institute{Kyung Hee University, Yongin, Republic of Korea \\
% \email{\{hyeonbae.kim, yhahn, st.kim\}@khu.ac.kr}\\}
\email{\{hyeonbae.kim, yhahn, st.kim\}@khu.ac.kr}\\}
\def\thefootnote{$\dag$}\footnotetext{Equal contribution; ${\ast}$~Corresponding author.}
\maketitle              % typeset the header of the contribution
\begin{abstract}
Recent advancements in deep neural networks have shown promise in aiding disease diagnosis and medical decision-making. 
However, ensuring transparent decision-making processes of AI models in compliance with regulations requires a comprehensive understanding of the model's internal workings.
However, previous methods heavily rely on expensive pixel-wise annotated datasets for interpreting the model, presenting a significant drawback in medical domains.
In this paper, we propose a novel medical neuron concept annotation method, named Mask-free Medical Model Interpretation (MAMMI), addresses these challenges. By using a vision-language model, our method relaxes the need for pixel-level masks for neuron concept annotation. MAMMI achieves superior performance compared to other interpretation methods, demonstrating its efficacy in providing rich representations for neurons in medical image analysis. Our experiments on a model trained on NIH chest X-rays validate the effectiveness of MAMMI, showcasing its potential for transparent clinical decision-making in the medical domain.
The code is available at \url{https://github.com/ailab-kyunghee/MAMMI}.

\keywords{Interpretability  \and Explainability \and Neuron-Concept annotation.}
\end{abstract}

\section{Introduction}
\label{intro}

Recent research has demonstrated considerable progress in employing deep neural networks (DNNs) within diverse medical domains, highlighting their potential for aiding disease diagnosis and medical decision-making. 
Nevertheless, when deploying AI models for real-world usage, it is important for models to provide sufficient justification for their decisions in compliance with the European Union’s General Data Protection Regulation (GDPR) law~\cite{transparency22,gdpr17}. 
This regulation emphasizes the importance of transparent decision-making processes. 
Therefore, ensuring transparent clinical decisions of AI models requires a comprehensive understanding of the internal workings of the model~\cite{dissecting18,medicalsurvey20}.

To improve understanding of the internal workings of the model (i.e., interpretability), various methods have been proposed~\cite{netdissect17,tcav18}. 
Wu~\textit{et al.} investigate the concept of neurons in the model by a subjective evaluation from radiologists. Network Dissection~\cite{netdissect17} offers a detailed explanation of the model's internal mechanisms by automatically identifying concepts for individual neurons leveraging pixel-wise segment mask annotated datasets like Broden. 
Khakzar~\textit{et al.} proposed Towards Semantic Interpretation (TSI)~\cite{semantic21} by applying Network Dissection to the medical domain leveraging pixel-wise segment mask annotated datasets. TSI is better than \cite{wu2018expert} by reducing the efforts for annotating concepts for every neuron by medical experts, which increases the annotation cost in proportion to the number of neurons. However it still heavily depends on pixel-wise annotated datasets, leading to heavy dataset collection costs~\cite{www24}. It is a significant drawback, especially in medical domains where pixel-level annotation is especially expensive.

To address this issue, recent studies in computer vision (CV) domain~\cite{www24,falcon23,clipdissect23} have introduced neuron concept annotation methods that leverage vision-language models, such as CLIP~\cite{clip21}. 
These methods utilize separate image datasets (probing set) and a set of text (concept set), offering a significant advantage in interpreting internal model representations without relying on pixel-wise segment mask annotated datasets.
However, adaptation of the neuron concept annotation methods to the medical field necessitates careful consideration of three critical factors. 
Firstly, it is imperative to curate a concept set that incorporates medical-specific information, ensuring relevance to the domain. Secondly, the class-imbalanced data distribution prevalent in medical datasets might harm the performance of neuron concept annotation methods by affecting neuron representation selection.
Lastly, it needs a vision-language model for the medical domain to ensure a suitable representation of medical-specific images and terms.

To address these challenges, we present a novel medical neuron concept annotation method, named \textbf{Ma}sk-free \textbf{M}edical \textbf{M}odel \textbf{I}nterpretation(MAMMI), which takes into account these key aspects in this study. 
Comprehensive experiments have been conducted on a model trained on chest X-rays (NIH14). Experimental results demonstrate that MAMMI achieves superior performance compared to other methods, showcasing its efficacy in providing rich representations for neurons in medical image analysis. Our contributions are as follows:
\begin{itemize}
    \item We proposed a novel medical neuron concept annotation method, MAMMI, to address the limitations of the previous neuron concept annotation methods. By leveraging vision-language models, MAMMI stands out as a mask-free method, eliminating the necessity for pixel-level datasets and thereby reducing the costs associated with collecting expensive pixel-level data.
    \item We explore methods for neuron-concept association in the medical field and design the method by accounting for medical-specific attributes that are often overlooked. These include factors such as class-imbalanced data distribution, which is particularly prevalent in medical datasets.
    \item Extensive experiments have been conducted to evaluate the effectiveness of the MAMMI. MAMMI demonstrated superior performance both qualitatively and quantitatively in a medical setting compared to other methods.
\end{itemize}

\section{Related Work}
\noindent \textbf{Concept Activation Vectors:}
Kim et al.~\cite{tcav18} proposed an interpretation method based on Concept Activation Vectors (CAVs). 
CAVs are defined as vectors orthogonal to a linear classifier that separate neuron activations between a positive example set and a negative non-example set for a user-defined concept.
TCAV (Testing with CAVs) measures the sensitivity of the model to all concepts for a given prediction, allowing an explanation of which concepts the model's decision is based on. 
Additionally, it has been demonstrated to be applicable in the medical domain by reporting results on retinal fundus images~\cite{fundus18}. 
However, CAVs have limitations on detailed and diverse automatic explanations of the model's internal workings.

\noindent \textbf{Network Dissection:}
Bau et al.~\cite{netdissect17} proposed Network Dissection, which matches concepts by measuring the Intersection over Union (IoU) score between pixel-wise segmentation masks and individual neuron (e.g., convolution filter) feature maps.
Khakzar et al.~\cite{semantic21} proposed TSI and demonstrated the utility of detecting concepts within neurons of a thoracic disease diagnosis model, showcasing its effectiveness in the medical domain.
However, the aforementioned methods require pixel-wise annotation due to their dependence on an image-concept-matched dataset like Broden~\cite{netdissect17} or NIH14~\cite{wang2017chestx} with bounding boxes, which incurs expensive pixel-level annotation costs for dataset collection.

\noindent \textbf{Vison-Language Model-based Methods:}
Vision-language models, such as CLIP~\cite{clip21} model, have been utilized for matching neuron concepts without the help of pixel-wise annotated datasets. Examples include FALCON~\cite{falcon23}, CLIP Dissect~\cite{clipdissect23}, and WWW~\cite{www24}.
FALCON~\cite{falcon23} selects high-activated examples and finds similar captions in a large-scale caption dataset~\cite{laion21}. This approach relies on the assumption that there are captions representing neurons, which requires a diverse large-scale caption dataset, including various captions. 
However, building such a dataset with diverse captions, especially in the medical domain where specialized knowledge is required, can be challenging.
CLIP Dissect~\cite{clipdissect23} matches concepts with the concept activation matrix, calculated based on the inner product of CLIP image feature and text features. And the neuron activations of the probing set. 
WWW~\cite{www24} matches concepts by calculating the adaptive cosine similarity in CLIP space between a fixed number of high-activated examples for each neuron and the concept set. 
\section{Method}
There are important consideration points for designing neuron concept annotation in the medical domain. 
For instance, concept sets consisting of general words may not properly represent the concepts that the model learns from the medical datasets.
In section~\ref{mc}, we introduce a way for constructing a concept set considering medical-specific information.
Also, in the medical dataset, the number of samples between classes can be different (i.e., class imbalance), which can be another challenging point for neuron concept annotation, in particular, for representative example selection.
In section~\ref{aes}, we propose adaptive neuron representative image selection for considering class-imbalance distribution data prevalent in medical datasets. 
In section~\ref{method:MAMMI}, we provide a description of the overall flow of MAMMI, incorporating the introduced elements.

\begin{figure*}[t]
    \centering
    \includegraphics[width=\textwidth]{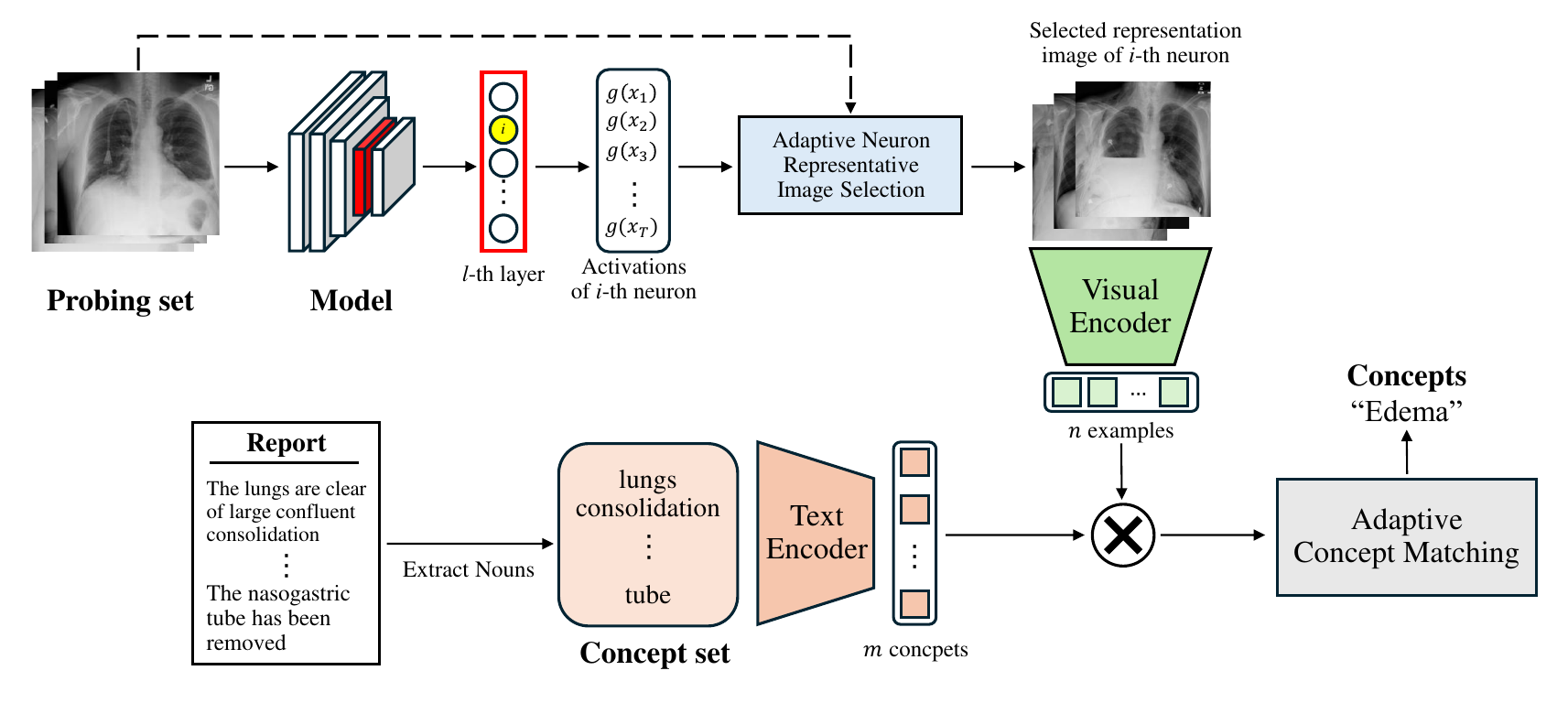}
    \caption{\small\textbf{Overview of neuron concept annotation method of MAMMI.} 
    The representative images of neurons for concept annotation are selected by images with activation values over the neuron-wise computed adaptive threshold from the probing data. The concept set is constructed by extracting nouns from the medical report. The representative images and the text concept set are projected into the CLIP model embedding space for concept identification by the adaptive concept matching module.
    }
    \label{fig:overall} 
\end{figure*}

\subsection{Constructing a Medical Concept Set}
\label{mc}

Our approach involves utilizing a dataset that has medical reports related to the domain of the trained target model.
Leveraging datasets that have target domain image-related reports gives great advantages in obtaining suitable concepts for interpreting the target model.
We extracted nouns in medical reports of the target domain to extract meaningful concepts for model interpretation.

\subsection{Adaptive Neuron Representative Image Selection}
\label{aes}
The class imbalance is one of the prevalent attributes of medical datasets.
Therefore, we propose an adaptive neuron representative image selection for each neuron representation.
Let neuron activations of target model $l-th$ layer $i-th$ neuron regarding concept set images denoted as $A^{(l,i)}$.
The adaptive threshold ($\tau$) for neuron representative images selection is computed as follows:
\begin{equation}\label{equ:thresh_exam}
    \tau^{(l,i)}_{example} = \max(A^{(l,i)}) - (1-\alpha) \cdot (\max(A^{(l,i)}) - \min(A^{(l,i)})),
\end{equation}
where $\alpha$ denotes the hyper-parameter for neuron representation selection sensitivity.
In Eq.~\ref{equ:thresh_exam}, for calculating the adaptation threshold of the \textit{i}-th neuron in the \textit{l}-th layer, we leverage the gap between the maximum sample activation ($\max(A^{(l,i)})$) and minimum sample activation ($\min(A^{(l,i)})$). 
Since each neuron has different activation patterns and ranges, leveraging the gap between maximum and minimum activation allows adaptive neurons representative image selection to consider the range of each neuron activation.
By adjusting the hyper-parameter $\alpha$, we can select suitable examples for neuron representation while considering the data imbalance.
We select $l-th$ layer $i-th$ neuron representative images $Imgs^{(l,i)} = \{A^{(l,i)}_k > \tau^{(l,i)}_{example}, (k = 1,2,...,p)\}$ where $p$ denotes the number of images in probing set.

\subsection{Overview of MAsk-free Medical Model Interpretation: MAMMI}
\label{method:MAMMI}
In this section, we provide an overview of the overall flow of MAMMI for medical neuron concept annotation. The concept identification follows the concept discovery module proposed in~\cite{www24}, and the overall flow is illustrated in Figure \ref{fig:overall}.

From the obtained images and concepts, the following steps are proceed for concept identification:
First, we calculate CLIP text features of concepts as $T=[t_1,t_2, ..., t_m]$ and CLIP visual features of a neuron to annotate (i.e., $l$-th layer $i$-th neuron) as $V^{(l,i)}=[v^{(l,i)}_{1},v^{(l,i)}_{2}, ... ,v^{(l,i)}_{n}]$ from $Imgs^{(l,i)}$. 
Here, $m$ denotes the total number of concepts, and $n$ represents the number of selected representing images of a neuron to annotate. 
Then, we measure the adaptive cosine similarity of the calculated features to compute the concept score $S$. 
The concept score of $j$-th concept regarding $l$-th layer $i$-th neuron $s^{(l,i)}_j$ is calculated as follows:
\begin{equation}\label{equ:Adaptive_sim}
    s^{(l,i)}_{j} = \frac{1}{n}\sum^{n}_{o = 1} \left\{cos(v^{(l,i)}_{o}, t_{j}) - cos(v^{(l,i)}_{o},t_{tem}) \right\},
\end{equation}
where $t_{tem}$ represents the CLIP text embedding of the base template.
If the calculated concept score $s^{(l,i)}_{j}$ is greater than the threshold $\theta_{concept}$, $j$-th concept is annotated as the concept for the corresponding neuron. The threshold $\theta_{concept}$ is calculated as $\theta_{concept} = \beta \times \max(s^{(l,i)})$.
$\beta$ denotes the ratio of selecting concepts. We used the same hyperparameter ($\beta$ = 0.95) as in~\cite{www24}.
\section{Experiments}
\subsection{Experimental Settings} The target model for interpretation is a DenseNet121 \cite{huang2017densely} architecture which is pre-trained with the MoCo training method and fine-tuned using the chest X-ray dataset~\cite{xiao2023delving}.
Additionally, for qualitative results, we employed ChestX-det~\cite{liu2020chestx} with bounding box annotations.
For quantitative evaluation, we performed it using the final layer neuron concepts and class labels as introduced in~\cite{clipdissect23,www24}. We calculated the similarity between the class label feature encoded by CLIP (ViT-B/16)~\cite{radford2021clip} and mpnet~\cite{song2020mpnet} and the neuron concepts. 
The F1-score indicates the exactness and flexibility of neuron concepts, while the hit-rate measures the extent to which neuron concepts correctly predict class labels. 

\begin{table}[t]
\caption{Ablation study on the concept set selection of MAMMI.}
\centering
\scalebox{0.85}{
\begin{tabular}{c|c|cc|cc}
\hline
\multirow{2}{*}{$D_{concept}$} & \multirow{2}{*}{\begin{tabular}[c]{@{}c@{}}\# of\\ concepts\end{tabular}} & \multicolumn{2}{c|}{Cosine Similarity} & \multirow{2}{*}{F1-score} & \multirow{2}{*}{Hit-rate}    
\\
 & & CLIP & mpnet & & 
 \\ \hline
20K~\cite{clipdissect23} & 20K  & 0.7749\tiny${\pm{0.016}}$ & 0.1045\tiny${\pm{0.016}}$ & 0.000   & 0.000 
\\
Wordnet nouns\cite{www24}  & 80K  & 0.7253\tiny${\pm{0.022}}$ & 0.1542\tiny${\pm{0.014}}$ & 0.000   & 0.000 
\\
MIMIC$_{Nouns}$ & 1361 & \textbf{0.8599}\tiny${\pm{0.024}}$ & \textbf{0.5290}\tiny${\pm{0.084}}$ & \textbf{0.2619}\tiny${\pm{0.113}}$ & \textbf{0.2857}\tiny${\pm{0.121}}$ 
\\ \hline
\end{tabular}
}
\label{tab:concept}
\end{table}

\subsection{Evaluation of Our Method}
\noindent \textbf{Ablation study on the Concept Set Selection of MAMMI:}
To evaluate the performance of the concept set considering medical-specific information, we compared various concept sets obtained from other sources.
In this experiment, we interpreted the target model fine-tuned on the NIH Chest X-ray14 (NIH14)~\cite{wang2017chestx}.
We collect concept set from MIMIC-CXR reports~\cite{johnson2019mimic} by extracting nouns from the datasets (i.e.,MIMIC$_{Nouns}$).
As a control group for MIMIC$_{Nouns}$, we compared two concept sets proposed in~\cite{clipdissect23} (20k) and proposed in~\cite{wordnet,www24} (80k) from the CV domain.

In Table~\ref{tab:concept}, the cosine similarity scores of the concept set from 20k and 80k sets have significantly decreased compared to the concept set of our method. Additionally, the F1 Score and hit rate, which evaluates whether annotated concepts match with labels or not, are both zero. 
This result indicates that by building a concept set with medical-specific information ($MIMIC_{Nouns}$), the model interpretability can be significantly improved.

\begin{table}[t]
\begin{minipage}{0.5\linewidth}
\centering
\caption{Ablation on the adaptive neuron representative image selection.}
\scalebox{0.75}{
\begin{tabular}{c|cc|c}
\hline

\multirow{2}{*}{\begin{tabular}[c]{@{}c@{}}Example\\ Selection\end{tabular}} & \multicolumn{2}{c|}{Cosine Similarity}  & Penultimate    \\
                                                                             & CLIP               & mpnet                         & Unique concept \\ \hline
w/o Adaptive & 0.8495\tiny${\pm{0.026}}$ & 0.5107\tiny${\pm{0.085}}$ & 65\\
Adaptive   & \textbf{0.8599\tiny${\pm{0.024}}$} & \textbf{0.5290\tiny${\pm{0.084}}$} & \textbf{96}\\ \hline
\end{tabular}
}
\label{tab:example-selection}
\end{minipage}
\hspace{10pt}
\begin{minipage}{0.5\linewidth}
\centering
\caption{Comparison of medical CLIP models in MAMMI.}
\scalebox{0.8}{
\begin{tabular}{c|ll}
\hline
\multirow{2}{*}{CLIP Model} & \multicolumn{2}{c}{Cosine Similarity} \\
  & \multicolumn{1}{c}{CLIP} & \multicolumn{1}{c}{mpnet} \\ \hline
CXR-RePaIR~\cite{cxrrepair21}  & 0.8334\tiny${\pm{0.018}}$ & 0.3716\tiny${\pm{0.070}}$ \\
MedKLIP~\cite{medclip22}     & 0.7704\tiny${\pm{0.014}}$ & 0.1780\tiny${\pm{0.022}}$ \\
MedCLIP~\cite{medclip22}     & \textbf{0.8599\tiny${\pm{0.024}}$} & \textbf{0.5290\tiny${\pm{0.084}}$} \\ \hline
\end{tabular}
 }
\label{tab:ablation-clip}
\end{minipage}
\end{table}

\begin{figure*}[t]
    \centering
    \includegraphics[width=0.9\textwidth]{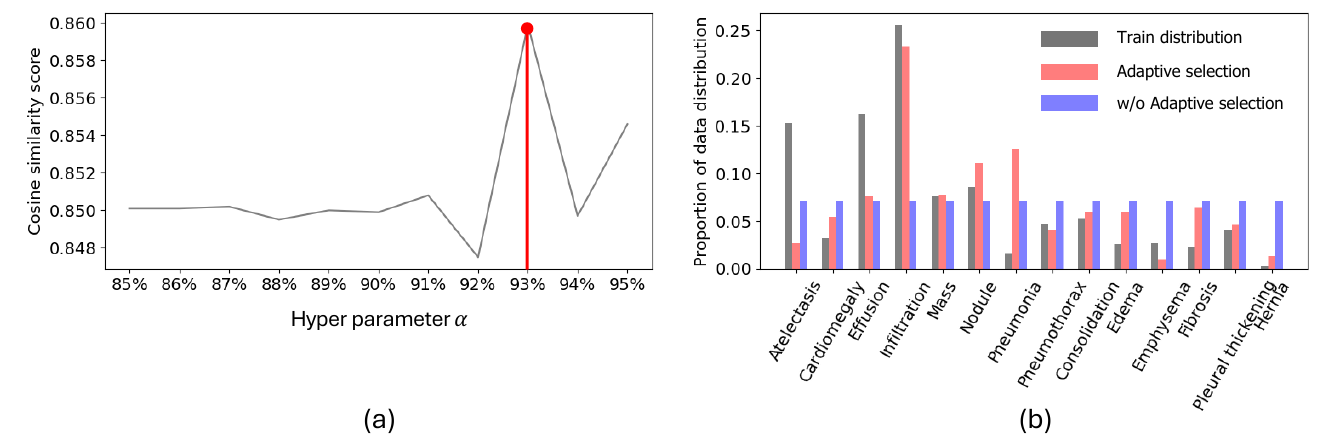}
    \caption{The data distribution of adaptive neuron representative image selection compared to the probing set. (a) Ablation study on the value of parameter $\alpha$. (b) Distribution regarding the number of selected representative images of neurons in the final layer and train set class distribution.
    }
    \label{fig:dist} 
\end{figure*}

\noindent \textbf{Ablation Study on Adaptive Neuron Representative Image Selection:}
In Table~\ref{tab:example-selection}, we present the results of an ablation study conducted under the same conditions for our proposed adaptive neuron representative image selection method and select a fixed number of examples (i.e., w/o adaptive neuron representative image selection).
The parameter $\alpha$ used in our adaptive neuron representative image selection is set to 93$\%$, 
We discovered an interesting finding when comparing the example distribution of adaptive neuron representative image selection with the data distribution of the probing set (NIH14).
Figure \ref{fig:dist} shows the proportion of class-wise data distribution in the probing set and the number of selected examples with and without adaptive neuron representative image selection in each neuron.
Compared to the method without adaptive selection, the distribution of adaptive neuron representative image selection shows a similar distribution to the train data distribution.
This result indicates that by leveraging neuron representative image selection, MAMMI can select suitable examples with less noisy representing images.
As a result, using adaptive neuron representative image selection showed marginal performance improvement.

\noindent \textbf{Ablation Study on Medical CLIP Model Selection:}
For better annotations for the neuron concepts of the target model, we conducted an ablation study by selecting the CLIP model trained on the chest X-ray dataset.
We compared three CLIP-based models, CXR-RePaIR~\cite{cxrrepair21}, MedKLIP~\cite{medclip22}, and MedCLIP~\cite{medclip22}.
Results are listed in Table \ref{tab:ablation-clip}.
Our experimental results showed that MedCLIP~\cite{medclip22} outperforms compared to other competitive baselines.

\subsection{Performance Evaluation with Other Methods}
\begin{figure*}[]
    \centering
    \includegraphics[width=0.98\textwidth]{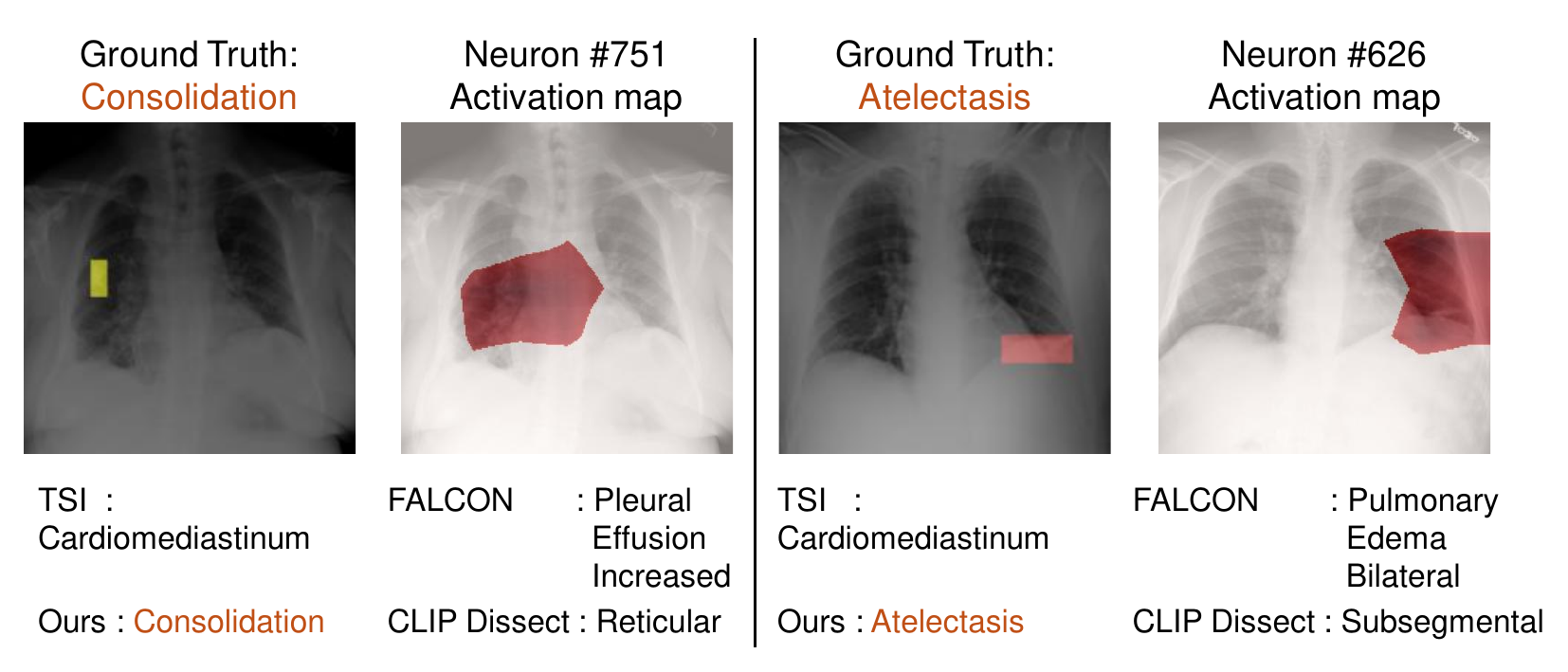}
    \caption{Qualitative result on the penultimate layer. We display two penultimate layer neurons selected by Shapley Value~\cite{khakzar2021neural,www24} as the most important penultimate neurons for identifying the respective ground truth label in the image.}
    \label{fig:qualitative} 
\end{figure*}

\noindent \textbf{Comparison with Mask-based Method:}
Table~\ref{tab:compare-mask} presents a performance comparison with the TSI~\cite{semantic21}, which utilizes mask annotation in the medical domain.
TSI used a concept set that consists of COVID-19 segmentation data (COVID-CXR)~\cite{COVID-CXR2020} and class labels from NIH Chest X-ray14 (NIH14)~\cite{wang2017chestx}.
MAMMI exhibited superior performance across all metrics, highlighting its effectiveness. 
Notably, TSI failed to accurately identify neuron concepts, as observed from F1-scores and Hit-rate scores.
For qualitative comparison, we display two penultimate layer neurons that are important for identifying ground truth labels with their respective activation regions.
Qualitative results in Figure~\ref{fig:qualitative}, MAMMI accurately identifying neuron concepts, even at the penultimate layer, while TSI~\cite{semantic21} failed to identify proper concepts.

\begin{table}[t]
\caption{Quantitative comparison with mask-based concept annotation method.}
\centering
\scalebox{0.9}{
\begin{tabular}{c|c|cc|cc}
\hline
\multirow{2}{*}{Method} & \multirow{2}{*}{$D_{concpet}$} & \multicolumn{2}{c|}{Cosine Similarity} & \multirow{2}{*}{F1-score} & \multirow{2}{*}{Hit-rate} \\
 &   & CLIP & mpnet   &       &     \\ \hline
TSI~\cite{semantic21}  & NIH14~\cite{wang2017chestx}+COVID-CXR~\cite{COVID-CXR2020}  & 0.7872\tiny${\pm{0.014}}$ & 0.2554\tiny${\pm{0.039}}$ & 0             & 0             \\
Ours & MIMIC$_{Nouns}$ & \textbf{0.8599\tiny${\pm{0.024}}$} & \textbf{0.5290\tiny${\pm{0.084}}$} & \textbf{0.2619\tiny${\pm{0.113}}$} & \textbf{0.2857\tiny${\pm{0.121}}$} \\ \hline
\end{tabular}
}
\label{tab:compare-mask}
\end{table}

\noindent \textbf{Comparison with Vision-Language Model-based Methods:}
Table \ref{tab:compare-cv} provides a performance comparison by implementing Vision-Language model-based methods from the CV domain into the medical setting. WWW is not explicitly mentioned as it serves as the baseline for MAMMI.
In the medical setting, we replace the caption dataset of FALCON~\cite{falcon23} with MIMIC-CXR medical reports~\cite{johnson2019mimic} for this experiment. Additionally, to ensure the identification of concepts in every neuron, the example selection threshold was modified to 0.25.
CLIP Dissect~\cite{clipdissect23} also adjusted into medical settings with the medical concept set(i.e.,MIMIC$_{Nouns}$) with MedCLIP\cite{medclip22}. Moreover, performance was evaluated using the top-1 neuron concept, ensuring that neuron concepts were annotated in every neuron.
As shown in Table~\ref{tab:compare-cv}, MAMMI achieved high scores across all metrics compared to other competitive baselines.
Also, qualitative results are presented in Figure~\ref{fig:qualitative}. As shown in the figure, MAMMI accurately identified concepts even at the penultimate layer, while other baselines~\cite{falcon23,clipdissect23} failed to identify proper concepts.

\begin{table}[t]
\caption{Quantitative comparison with other concept annotation methods.}
\centering
\scalebox{0.9}{
\begin{tabular}{c|c|cc|cc}
\hline
\multirow{2}{*}{Method} & \multirow{2}{*}{D\_concept} & \multicolumn{2}{c|}{Cosine Similarity} & \multirow{2}{*}{F1-score} & \multirow{2}{*}{Hit-rate} \\
 & & CLIP & mpnet &  & \\ \hline
FALCON$_{M}$       & MIMIC report~\cite{johnson2019mimic} & 0.8054\tiny${\pm{0.012}}$ & 0.2420\tiny${\pm{0.031}}$ & 0.0635\tiny${\pm{0.027}}$ & \textbf{0.2857\tiny${\pm{0.121}}$} \\
CLIP Dissect$_{M}$ & MIMIC$_{Nouns}$ & 0.8205\tiny${\pm{0.024}}$ & 0.3700\tiny${\pm{0.083}}$ & 0.0714\tiny${\pm{0.069}}$ & 0.0714\tiny${\pm{0.069}}$ \\
Ours               & MIMIC$_{Nouns}$ & \textbf{0.8599\tiny${\pm{0.024}}$} & \textbf{0.5290\tiny${\pm{0.084}}$} & \textbf{0.2619\tiny${\pm{0.113}}$} & \textbf{0.2857\tiny${\pm{0.121}}$} \\ \hline
\end{tabular}
} 
\label{tab:compare-cv}
\end{table}

\section{Conclusion}
In this study, we introduce a novel approach for interpreting medical vision models via neuron concept association named MAMMI, specifically tailored for the medical domain. 
Recognizing the challenges associated with constructing new datasets, we devised a concept set considering medical-specific information.
Also, we employed an adaptive neuron representative image selection method to overcome the imbalanced data distribution of common attributes in medical datasets. 
Our extensive experiments demonstrated that MAMMI outperforms other approaches in annotating a wide variety of concepts to neurons in medical models.
Furthermore, our method addresses a limitation of existing approaches, such as TSI, which requires expensive mask-annotated datasets. 
\begin{credits}
\subsubsection{\ackname} This work was supported in part by the National Research Foundation of Korea(NRF) grant funded by the Korea government(MSIT) (No.RS-2024-00334321) and by the Institute of Information and Communications Technology Planning and Evaluation (IITP) Grant funded by the Korea Government (MSIT) under Grant 2022-0-00078, Grant IITP-2024-RS-2023-00258649, and Grant RS-2022-00155911.

\subsubsection{\discintname}
The authors have no competing interests to declare that are relevant to the content of this article. 
\end{credits}

\bibliographystyle{splncs04}
\bibliography{ref}

\end{document}